\documentclass[3p,twocolumn]{elsarticle}

\usepackage{lineno,hyperref}
\usepackage{array}
\usepackage{textcomp}
\usepackage{multirow}
\usepackage{amstext}
\usepackage{amssymb}
\usepackage{graphicx}
\usepackage{esint}
\usepackage{color}
\usepackage{soul}
\usepackage[table]{xcolor}
\usepackage{url}
\usepackage{siunitx}
\usepackage{subfig}

\journal{Journal of Medical Image Analysis}

\newcommand{\DR}{D-ROB}
\newcommand{\DC}{D-CONV}

\newcommand{\DSR}{\DR-SEG}
\newcommand{\DSRAll}{\DSR\textsubscript{All}}

\newcommand{\DSC}{\DC-SEG}
\newcommand{\DSCAll}{\DSC\textsubscript{All}}
\newcommand{\DSCBlood}{\DSC\textsubscript{Blood}}
\newcommand{\DSCMesh}{\DSC\textsubscript{Mesh}}
\newcommand{\DSCOverlap}{\DSC\textsubscript{Overlap}}
\newcommand{\DSCSmoke}{\DSC\textsubscript{Smoke}}

\newcommand{\DTR}{\DR-TRA}
\newcommand{\DTRAll}{\DTR\textsubscript{All}}
\newcommand{\DTRMultiple}{\DTR\textsubscript{Multiple}}

\newcommand{\DTC}{\DC-TRA}
\newcommand{\DTCAll}{\DTC\textsubscript{All}}
\newcommand{\DTCBlood}{\DTC\textsubscript{Blood}}
\newcommand{\DTCMultiple}{\DTC\textsubscript{Multiple}}
\newcommand{\DTCObjects}{\DTC\textsubscript{Objects}}
\newcommand{\DTCOcclusion}{\DTC\textsubscript{Occlusion}}
\newcommand{\DTCSmoke}{\DTC\textsubscript{Smoke}}

\newcolumntype{L}[1]{>{\raggedright\let\newline\\\arraybackslash\hspace{0pt}}m{#1}}
\graphicspath{{pics/}}

\begin{document}

\begin{frontmatter}
\title{Comparative evaluation of instrument segmentation and tracking methods in minimally invasive surgery}
\author[NCT]{Sebastian Bodenstedt\corref{cor2}}
\ead{sebastian.bodenstedt@nct-dresden.de}
\author[IS]{Max Allan}
\author[UGA]{Anthony Agustinos}
\author[UCL]{Xiaofei Du}
\author[UCL]{Luis Garcia-Peraza-Herrera}
\author[HD]{Hannes Kenngott}
\author[ARTOG]{Thomas Kurmann}
\author[HD]{Beat M{\"u}ller-Stich}
\author[UCL]{Sebastien Ourselin}
\author[JHU]{Daniil Pakhomov}
\author[ARTOG]{Raphael Sznitman}
\author[IAR]{Marvin Teichmann}
\author[IAR]{Martin Thoma}
\author[UCL]{Tom Vercauteren}
\author[UGA]{Sandrine Voros}
\author[HD]{Martin Wagner}
\author[IAR]{Pamela Wochner}
\author[DKFZ]{Lena Maier-Hein}
\author[UCL]{Danail Stoyanov\fnref{cor1}}
\author[NCT]{Stefanie Speidel\corref{cor2}\fnref{cor1}}
\ead{stefanie.speidel@nct-dresden.de}

\fntext[cor1]{Contributed equally}
\cortext[cor2]{Corresponding authors}
\address[NCT]{Department for Translational Surgical Oncology, National Center for Tumor Diseases (NCT), Dresden, Germany}
\address[IS]{Intuitive Surgical, Sunnyvale, USA}
\address[UGA]{Universit{\'e} Grenoble Alpes, CNRS, CHU Grenoble Alpes, Grenoble INP, TIMC-IMAG, Grenoble, France}
\address[UCL]{Centre for Medical Image Computing, University College London, UK}
\address[HD]{Department of General, Visceral and Transplant Surgery, University of Heidelberg, Heidelberg, Germany}
\address[ARTOG]{ARTORG Center, University of Bern, Bern, Switzerland}
\address[JHU]{Computer Aided Medical Procedures, Johns Hopkins University, Baltimore, USA}
\address[IAR]{Institute for Anthropomatics and Robotics, Karlsruhe Institute of Technology (KIT), Karlsruhe, Germany}
\address[DKFZ]{Computer Assisted Medical Interventions, German Cancer Research Center (DKFZ), Heidelberg, Germany}

\begin{abstract}
Intraoperative segmentation and tracking of minimally invasive instruments is a prerequisite for computer- and robotic-assisted surgery. 
Since additional hardware like tracking systems or the robot encoders are cumbersome and lack accuracy, surgical vision is evolving as promising techniques to segment and track the instruments using only the endoscopic images. However, what is missing so far are common image data sets for consistent evaluation and benchmarking of algorithms against each other.
The paper presents a comparative validation study of different vision-based methods for instrument segmentation and tracking in the context of robotic as well as conventional laparoscopic surgery.
The contribution of the paper is twofold: we introduce a comprehensive validation data set that was provided to the study participants and present the results of the comparative validation study.
Based on the results of the validation study, we arrive at the conclusion that modern deep learning approaches outperform other methods in instrument segmentation tasks, but the results are still not perfect.
Furthermore, we show that merging results from different methods actually significantly increases accuracy in comparison to the best stand-alone method.
On the other hand, the results of the instrument tracking task show that this is still an open challenge, especially during challenging scenarios in conventional laparoscopic surgery.
\end{abstract}

\begin{keyword}
Endoscopic Vision Challenge \sep Instrument segmentation \sep Instrument tracking \sep Laparoscopic surgery \sep Robot-assisted surgery
\end{keyword}

\end{frontmatter}


\section{Introduction}
Minimally invasive surgery using cameras to observe the internal anatomy is the preferred approach for many surgical procedures.
This technique reduces the operative trauma, speeds recovery and shortens hospitalization. 
However, such operations are highly complex and the surgeon must deal with  a difficult hand-eye coordination, a restricted mobility and a narrow field of view \cite{Bernhardt201766}.
Surgeons capabilities can be enhanced with  computer- and robotic-assisted surgical systems \cite{cleary2010image}.
Such systems provide additional patient-specific information during surgery, e.g. by visualizing hidden risk and target structures based on preoperative planning data.
Intraoperative localization of minimally invasive instruments is a prerequisite for such systems.
The pose of the instrument is crucial for e.g. measuring the distance to risk structures \cite{Bernhardt201766}, automation of surgical skills \cite{chen2016virtual} or assessing the skill level of a surgeon \cite{vedula2016objective}. Since additional hardware like tracking systems, instrument markers or the robot encoders are cumbersome and lack accuracy, surgical vision is evolving as promising technique to localize the instruments using solely the endoscopic images.
Image-based localization can be split into segmentation and tracking of the instruments in the endoscopic view.
	
Image-based instrument segmentation and tracking has received increased attention in different minimally invasive scenarios. A recent paper by Bouget et al. provides an in-depth review of different instrument detection and tracking algorithms \cite{bouget2017}. However, what is missing so far are common image data sets for consistent evaluation and benchmarking of algorithms against each other. 
	
\begin{figure*}[t]
	\centering
	\includegraphics[width=0.9\textwidth]{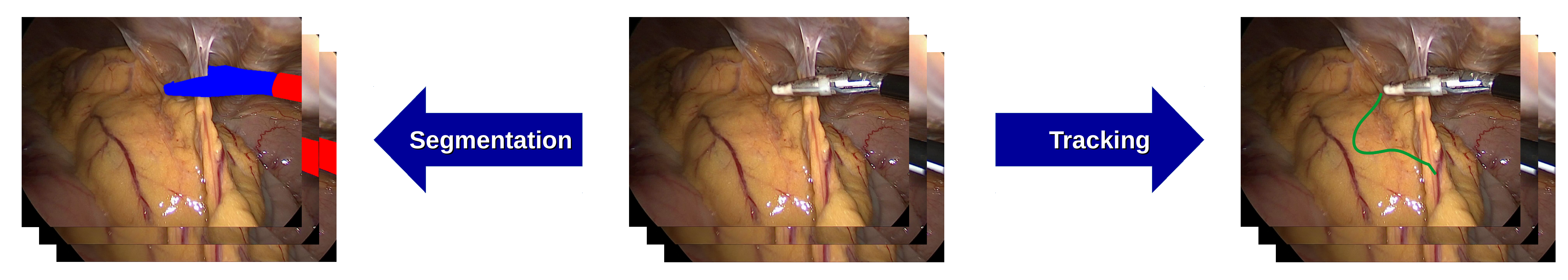}
	\caption{The two task during the \textit{Instrument segmentation and tracking} sub-challenge.}
	\label{fig:overview}
\end{figure*}

In this paper, we present a comparative validation of different vision-based state-of-the-art methods for instrument segmentation as well as tracking in the context of minimally invasive surgery (figure \ref{fig:overview}). The data used is based on the data of the sub-challenge \textit{Instrument segmentation and tracking} \url{https://endovissub-instrument.grand-challenge.org/}, part of the \textit{Endoscopic Vision Challenge} (\url{http://endovis.grand-challenge.org}), at the international conference \textit{Medical Image Computing and Computer Assisted Intervention}.
	
The contribution of the paper is twofold: we introduce a comprehensive validation data set that was provided to the study participants and present the results of the comparative validation.
	
Two important surgical application scenarios were identified: robotic as well a conventional laparoscopic surgery. Both scenarios face different challenges considering the instruments. We have articulated instruments in robotic and rigid instruments in conventional laparoscopic surgery. 
Corresponding validation data was generated for both scenarios and consisted of endoscopic \textit{ex-vivo} images with articulated robotic as well as \textit{in-vivo} images with rigid laparoscopic instruments. The data was split into training and test data for the segmentation as well as the tracking task.
All data used in this paper is publicly available at \url{http://open-cas.org/?q=node/31}.

The paper is organized as follows. Section \ref{sec:methods_seg} and \ref{sec:methods_tra} briefly review the segmentation and tracking methods that participated in the study. Section \ref{sec:valdata} describes the validation data sets followed by the comparative validation of both tasks in section \ref{sec:study_seg} and \ref{sec:study_track}. We present the study results in section \ref{sec:results} and discuss our findings in section \ref{sec:discussion}. Finally, section \ref{sec:conclusion} provides a conclusion. 

\section{Instrument segmentation methods} 
\label{sec:methods_seg}
This section briefly reviews the basic working principle of the different segmentation methods that participated in the comparison study. Most of the approaches investigated were either based on random forests (RF) (\ref{subsec:segkitrf}, \ref{subsec:seguclrf}) or convolutional neural networks (CNN) techniques (\ref{subsec:segjhu}, \ref{subsec:segkitcnn}, \ref{subsec:segub}, \ref{subsec:seguclcnn}), except \ref{subsec:seguga}.

\subsection{SEG-JHU}
\label{subsec:segjhu}
The approach \cite{pakhomov2017deep} is based on the CNN described in \cite{xie2015holistically}.
The architecture, which was originally designed for edge detection problems, is similar to FCN-8s (fully convolutional network) as described in \cite{Long_2015_CVPR} and is adopted for the task of binary segmentation. 
A Deep Supervision approach is used for training. 
Based on experiments, the proposed architecture shows comparable or better performance on surgical instruments segmentation across all datasets compared to FCN-8s network \cite{Long_2015_CVPR}. 

\subsection{SEG-KIT-CNN}
\label{subsec:segkitcnn}
The first approach from Karlsruhe Institute of Technology (KIT) uses a FCNs \cite{Long_2015_CVPR} for semantic segmentation of the instruments and background. This type of neural network consists of two parts: An encoder, which is trained on an image classification task, and a decoder, which upsamples the result to the required original resolution. The encoder is initialized with VGG-16~\cite{Simonyan2014}, the decoder is a single layer of two upsampling filters of the size $64 \times 64$. There are two upsampling filters as two classes have to be recognized. The FCN is trained end-to-end with the mean softmax cross entropy between logits and labels as training objective. Each weight is regularized with L2 loss.

\subsection{SEG-KIT-RF}
\label{subsec:segkitrf}
The second approach from the KIT segments the instruments in the endoscopic images based on a feature vector for each pixel consisting of values from multiple color spaces, such as HSV, RGB, LAB and Opponent, and gradient information \cite{Bodenstedt2016a}. Using the supplied masks, a RF classifier to distinguish instrument pixels from background pixels using OpenCV is trained. For real-time online segmentation, a GPU-based RF is used. After the classification step, contours are located and fused if they are in close proximity to one another. Contours whose size lies over a certain threshold are then returned as instrument candidates.

\subsection{SEG-UB}
\label{subsec:segub}
The method developed by the University of Bern (UB) uses a CNN for classifying pixels as instrument or background. An AlexNet \cite{NIPS2012_4824} CNN architecture is used with a two neuron output in the last fully connected layer. During training, real time data augmentation is applied. The patches sizing  $227 \times 227$ pixels are randomly rotated, scaled, mirrored and illumination adjusted. Every training batch had 70\% background patches and 30\% instrument patches. As the computational requirements of CNNs are relatively large, only a raster of pixels with a stride of 7 pixels in both image axes is used. After classification the pixels are grouped using DBSCAN clustering \cite{Ester96}. Using the information from the perimeter of the clustered regions, an alpha shape \cite{edelsbrunner1983} is created. The region that is enclosed by the alpha shape is used as the final segmentation result. A combination of the Caffe Deep Learning Framework \cite{jia2014caffe} and Matlab 
were used to implement the described method.

\subsection{SEG-UCL-CNN}
\label{subsec:seguclcnn}
The first approach from the University College London (UCL) is a real-time segmentation method that combines deep learning and optical flow tracking \cite{Herrera17}.
Fully Convolutional Networks (FCNs) are able to produce accurate segmentations of deformable surgical instruments.
However, even with state-of-the-art hardware, the inference time of FCNs exceeds the framerate of the surgical endoscopic videos.
SEG-UCL-CNN leverages the fact that optical flow can track displacements of surgical tools at high speed to propagate FCN segmentations in real time. 
A parallel pipeline is used where a FCN runs in asynchronous fashion, segmenting only a small proportion of the frames. New frames are used as input for the FCN inference process only when the neural network is idle. The output of the FCN is stored in synchronized memory. In parallel, for 
every frame of the video, previously detected keypoints from stored frames are matched to detected keypoints in the current to-be-segmented frame and used to estimate an affine transformation with RANSAC. The segmentation mask of the stored frame is warped with the affine transformation estimated to produce the final segmentation for the frame with less than one inter-frame delay. 

\subsection{SEG-UCL-RF}
\label{subsec:seguclrf}
The second approach of Allan et al. from UCL is based on a RF classification \cite{Allan2013a}. They use variable importance to select 4 color features: Hue, Saturation and Opponent 1 and 2, which the experiments demonstrated provided the most discriminative power. The RF is trained on these features to classify the pixels in the test set with no post-processing stages. In all of the experiments, the OpenCV CPU RF implementation was used.

\subsection{SEG-UGA}
\label{subsec:seguga}
The method by Agustinos et al. from the Universit{\'e} Grenoble Alpes (UGA) uses color and shape information \cite{Agustinos2016} to segment the instruments. Based on the CIELab color space, a grayscale image composed of the a and b channels, corresponding to the chromaticity $C_{ab}$, is computed. Afterwards, a postprocessing step consisting of binarization with an automatic Otsu thresholding and a skeletonization using a simple distance transform \cite{felzenszwalb2004} followed by an erosion is performed. A contour detection algorithm \cite{suzuki1985} is then used to extract the extreme outer contour of each region as an oriented bounding box. Bounding boxes which do not satisfy a specific shape constraint (width/length ratio not inferior to 2) are eliminated. For each candidate, a Frangi filter \cite{frangi1998} and a Hough transform is used to highlight the instrument edges inside the box. False candidates are eliminated based on the relative orientation and position of the detected lines.


\section{Instrument tracking methods} 
\label{sec:methods_tra}

\begin{figure*}[t]
	\centering
	\subfloat[]{\includegraphics[width=0.15\textwidth]{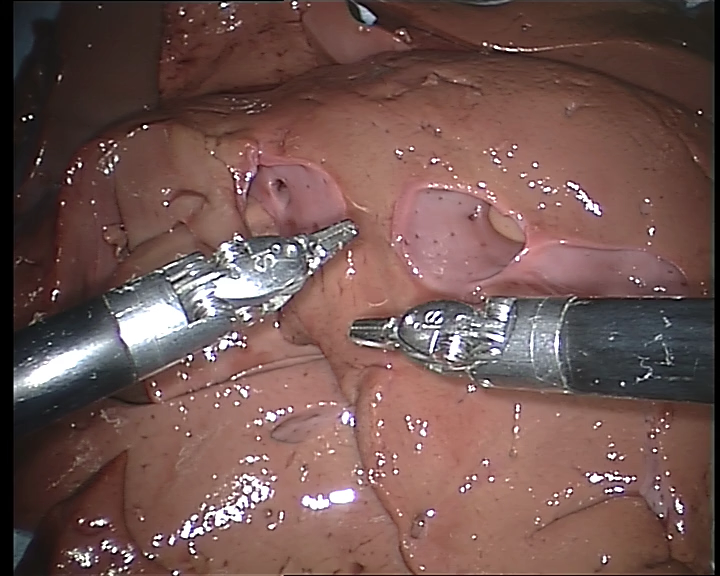}}\hfill
  \subfloat[]{\includegraphics[width=0.16\textwidth]{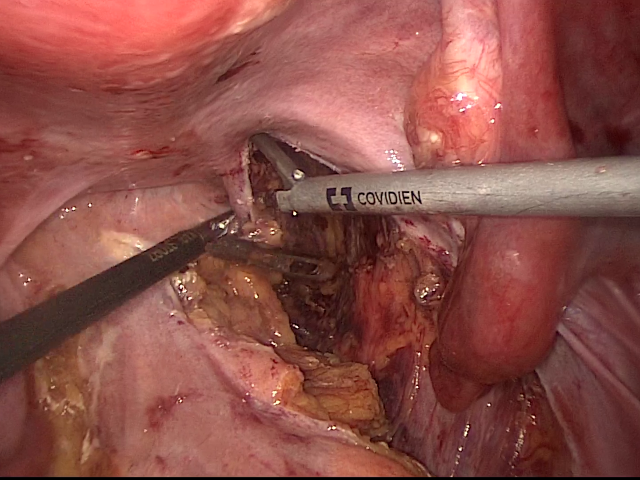}}\hfill
  \subfloat[]{\includegraphics[width=0.16\textwidth]{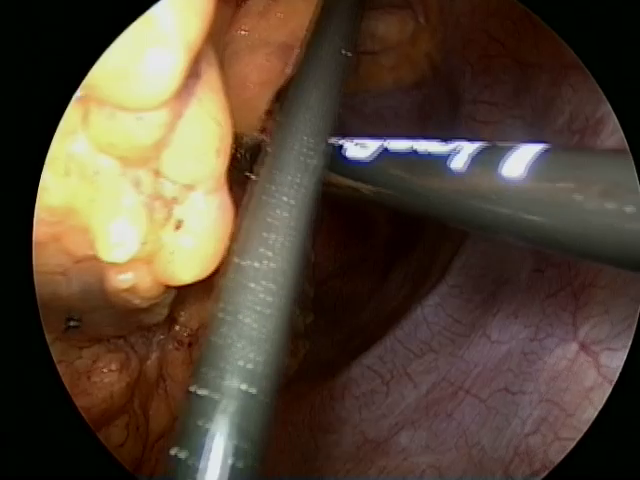}}\hfill
	\subfloat[]{\includegraphics[width=0.16\textwidth]{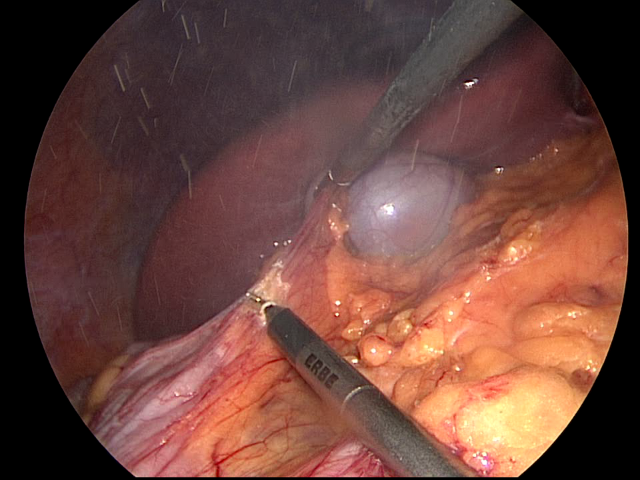}}\hfill
	\subfloat[]{\includegraphics[width=0.16\textwidth]{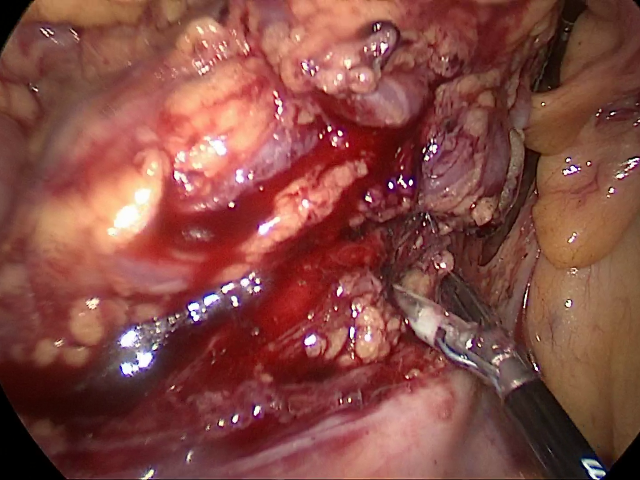}}\hfill
	\subfloat[]{\includegraphics[width=0.16\textwidth]{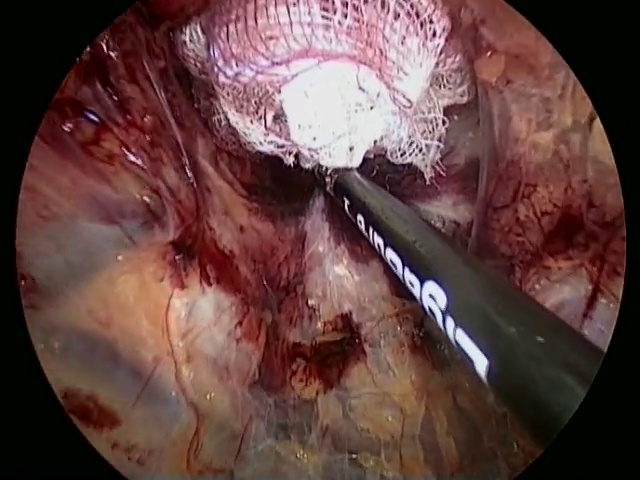}}
	\caption{Validation data set examples. (a) Robotic instruments. (b) Conventional instruments. (c)-(f): Challenges in conventional data set: overlapping instruments (c), smoke (d), bleeding (e), mesh (f).}
	\label{fig:seg-valdata}
\end{figure*}

This section briefly reviews the basic working principle of the different tracking methods that participated in the comparison study. The following approaches are based on the segmentation methods that were described in the previous section. 

\subsection{TRA-KIT}
\label{subsec:trakit}
The approach from KIT utilizes the segmentation method outlined above \ref{subsec:segkitrf} to initialize the tracking. For each detected contour, a principle component analysis is performed to find the axis and center of the tool. The furthest point from the image border, which lies on both the axis and the contour, is used as instrument tip. Furthermore, a bounding box around each detected instrument is calculated.
On each bounding box, features to track are detected by randomly sampling edges located with the Canny edge detector. Using Lucas-Kanade optical flow a vector describing the motion of each feature to its position in the next frame is computed. Afterwards a consensus over all directional vectors to locate the position of the instrument tip in the next frame is formed.

\subsection{TRA-UCL-MOD}
\label{subsec:trauclmod}
The method proposed by Allan et al. \cite{Allan2015} from UCL tracks the 2D tip location and instrument orientation by projecting the output of a 3D tracker. This tracker is based on aligning a CAD model projection with the output of the RF described in \ref{subsec:seguclrf} where the 3D pose is recovered as the model transformation which generates the best projection. This is formulated within a multi-region level set framework where the estimates with frame-to-frame optical flow tracking are combined.

\subsection{TRA-UCL-OL}
\label{subsec:trauclol}
The second approach from UCL \cite{Du2015} treats the tracking problem as a classification task, and the object model is updated over time using online learning techniques.
However, these methods are prone to include background information in the object appearance or lack the ability to estimate scale changes.
To address this problem, a Patch-based Adaptive Weighting with Segmentation and Scale (PAWSS) is used.
A simple colour-based segmentation model is then used to suppress the background information
Furthermore multi-scale samples are extracted online, which allows the tracker to handle incremental and abrupt scale variations between frames.
The method is based on online learning, it only needs bounding box initialization of the object on the first frame, then tracks the object for the whole sequence. 
The assumption of this kind of method is that the object is in view through all the sequence, so it can not handle out-of-view situation. 

\subsection{TRA-UGA}
\label{subsec:trauga}
The tracking approach from UGA \cite{Agustinos2016} assumes that an instrument does not undergo large displacements between two successive images. In the inital step (first image), the instruments are located as described in section \ref{subsec:seguga}. In the following images, the candidate bounding boxes are detected, but the instrument search is refined only inside the bounding box best compatible with the position/orientation of the instrument in the previous image. Knowing the border of an instrument, the position of its tip in the Frangi image along its central axis is detected. The pixel along this line with maximum grey level in the Frangi image is considered as the tip.



\section{Validation data}
\label{sec:valdata}

To compare the different segmentation and tracking methods from the participants,
we identified two important surgical application scenarios that capture a wide instrument type and background variety generally encountered in laparoscopic images. The first scenario is in the context of robotic laparoscopic surgery and includes articulated instruments, the second scenario deals with rigid instruments in conventional laparoscopic surgery. Corresponding image validation data was generated for both scenarios covering the different challenges considering the instruments (figure \ref{fig:seg-valdata}).

\subsection{Robotic laparoscopic instruments}
The articulated robotic instrument data set \textbf{\DR} originates from \textit{ex-vivo} 2D recordings using the da Vinci surgical system. In total, 6 videos from ex-vivo studies with varying background including a Large Needle Driver and Curved Scissors with $720 \times 576$ resolution were provided (table \ref{tab:valdata_overview}). The instruments show typical poses and articulation in robotic surgery including occlusion in any sequence, though artifacts such as smoke and bleeding were not included in the data. All data was provided in video files. Task-specific reference data includes annotated masks for different instrument parts as well as the 2D instrument pose. 

\subsection{Conventional laparoscopic instruments}
The conventional laparoscopic instrument data set \textbf{\DC} consists of 6 different \textit{in-vivo} 2D recordings from complete laparoscopic colorectal surgeries with $640 \times 480$ resolution (table \ref{tab:valdata_overview}). The images reflect typical challenges in endoscopic vision like overlapping instruments, smoke, bleeding and external materials like meshes. In total, the set contains seven different instruments that occur typically in laparoscopic surgeries  including hook, atraumatic grasper, ligasure, stapler, scissor and scalpel. From these 6 recordings, we extracted single frames for the segmentation task as well as video sequences for tracking. As reference data, annotated masks containing pixel-wise labels of different instrument parts as well as the 2D instrument pose was provided depending on the task.

\begin{table*}[htb]
\centering
\small
\begin{tabular}{|l||l|l|l|l|l|}
\hline
& Type &  \# of & $\oslash$ Size & Frame & \# types of  \\ 
& & videos& & size & instruments \\ 
\hline
\textbf{\DR} & ex-vivo & 6 & 1min & $720\times576$	&	2\\
\hline
\textbf{\DC} & in-vivo & 6 & 197min	& $640\times480$	&	7\\
\hline

\end{tabular}
\caption{Validation data for robotic/conventional laparoscopic segmentation/tracking.} 
\label{tab:valdata_overview} \end{table*} 


\section{Comparative study: Instrument segmentation}
\label{sec:study_seg}

The following sections provide a description of the training and test data for the segmentation task, the reference method and the validation criterion.


\subsection{Training and test data}
\label{ssec:train_seg}

\begin{table*}[htb]
\centering
\small
\begin{tabular}{|l||l|l|l|l|l|}
\hline
& \#Seq & Size     & \multicolumn{3}{ c| }{Data Split}    \\ 
							&       &          & Seq ID & Train & Test    \\ \hline
\multirow{2}{*}{\textbf{\DSR}} &    \multirow{2}{*}{6}      & \multirow{2}{*}{60s}   & 1--4	&		75\%		&25\%				\\
				&  			&                     & 5--6 &   - & 100\% \\ \hline
\multirow{2}{*}{\textbf{\DSC}} &    \multirow{2}{*}{6}      & \multirow{2}{*}{50 frames}   & 1--4	&		80	\%	&20\%				\\
				&  			&                     & 5--6 &   - & 100\% \\ \hline
\end{tabular}
\caption{Training and test data for robotic/conventional instrument segmentation.} 
\label{tab:valdata_seg} \end{table*}


As described in the previous paragraph \ref{sec:valdata}, the validation data originated from recordings of \textit{ex-vivo} robotic (\DR) and \textit{in-vivo} conventional (\DC) laparoscopic surgeries. 

\begin{figure}
	\centering
	\subfloat{\includegraphics[width=0.33\columnwidth]{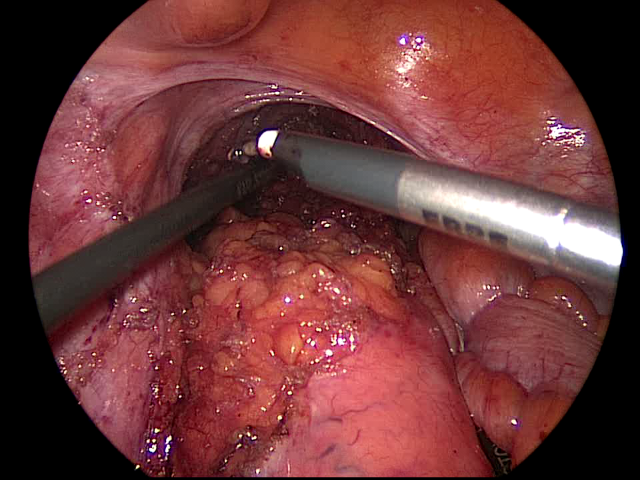}}\hfill
  \subfloat{\includegraphics[width=0.33\columnwidth]{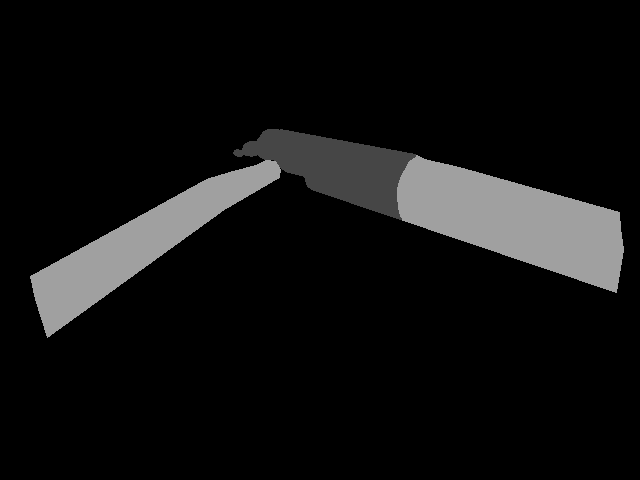}}\hfill
  \subfloat{\includegraphics[width=0.33\columnwidth]{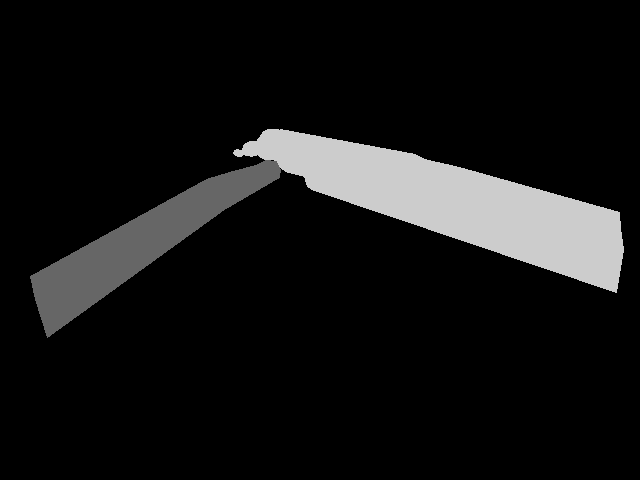}}
	\caption{Training data example including RGB image, mask for instrument parts and mask for instrument types}
	\label{fig:seg-traindata}
\end{figure}

The training data includes RGB frames as well as two types of annotated masks as reference data (figure \ref{fig:seg-traindata}). The first mask contains pixel-wise labels for the different instrument parts (shaft/manipulator) and the background. Furthermore, an additional mask with labels for each instrument is available if more than one instrument is in the scene. For the test data no masks are provided. A detailed overview including training and test data is given in table \ref{tab:valdata_seg}.

For the one minute robotic sequences, an annotation is provided for every frame.
Since the frames for conventional segmentation originate from complete surgeries, single frames had to be selected in a standardized manner. For the selection, each \textit{in-vivo} recording was divided into 50 segments where 10 frames from each segment where randomly extracted. From these 10 frames, one was manually selected to guarantee a certain image quality. This resulted in a validation set of 50 images per recording. Furthermore, the validation data set for conventional laparoscopic surgery was categorized according to challenging situations including overlapping instruments (\DSCOverlap), bleeding (\DSCBlood), smoke (\DSCSmoke) and meshes (\DSCMesh) in the scene and the validation measures calculated separately for the subsets. These challenges were taken into consideration while dividing the validation set in training and testing data. 

Participants were advised to use the training data in a leave-one-sequence-out fashion. It was not allowed to use the same sequence in the training set when testing the additional images for each of the sequences (1-4) provided for training. For the new sequences (5-6) the whole training data could be used.
Participants uploaded pixel-wise instrument segmentation results for each frame in the test data set. The results were in the form of binary masks, indicating instrument or background.  

\subsection{Reference method and associated error}
\label{ssec:ref_seg}
Reference data is provided as annotated masks, which were generated differently according to the surgical application scenario. Although the reference annotation aims to be as accurate as possible, errors are associated with the generation and are discussed accordingly.

\subsubsection{Robotic laparoscopic instruments}
The pixel-wise labeling was generated through backprojection of a 3D CAD model with hand corrected robotic kinematics. The CAD model used per-vertex material properties, which allowed the mask values for both the shaft and metal head to be written directly to the corresponding pixels. As these values were provided in video files, errors due to video compression could arise and the values had to be thresholded to the nearest expected value.

\subsubsection{Conventional laparoscopic instruments}
The reference annotation was generated via crowdsourcing. In a first step, the Pallas Ludens crowdsourcing platform (Pallas Ludens GmbH, Heidelberg, Germany) was used to segment all endoscopic instruments using a similar approach as in \cite{maier2014a}. Next, human observers (non-experts) went through all annotations and made manual corrections if necessary. Finally, the challenge organizers double-checked the quality of all annotations.

\subsection{Validation criteria}
\label{ssec:vc}
To evaluate the quality of a submitted segmentation result, several measures were calculated between the reference (R) and the predicted segmentation (P) of a given method provided as binary mask. The criteria used included typical classification metrics like precision, recall and accuracy as well as the dice similarity coefficient (DSC) \cite{powers2011evaluation}. The DSC is a typical measure to evaluate the segmentation result by calculating the overlap of the predicted (P) and reference (R) segmentation. It is defined as: 

\begin{equation}
DSC = \frac{2\left|P \cap R\right|}{\left|P\right| + \left|R\right|}
\end{equation}

In the case of a binary segmentation, the DSC is identical to the F score, which is a combination of precision and recall.

\section{Comparative study: Instrument tracking}
\label{sec:study_track}

The following sections provide a description of the training and test data for the tracking task, the reference method and the validation criterion.

\subsection{Training and test data}
\label{ssec:train_track}

\begin{table*}[htb]
\centering
\small
\begin{tabular}{|l||l|l|l|l|l|l|}
\hline
& \#Seq & Size  & Mask  & \multicolumn{3}{ c| }{Data Split}    \\ 
&       &       &   & Seq ID & Train & Test    \\ \hline
\multirow{2}{*}{\textbf{\DTR}} &    \multirow{2}{*}{6}      & \multirow{2}{*}{60s}   &  every &1--4	&		75\%		&25\%				\\
				&  			&          &     frame      & 5--6 &   - & 100\% \\ \hline

\multirow{2}{*}{\textbf{\DTC}} &    \multirow{2}{*}{6}      & \multirow{2}{*}{60s}   & \multirow{2}{*}{1 fps}& 1--4	&	75\%	&25\%				\\
				&  			&        &             & 5--6 &   - & 100\% \\ \hline
\end{tabular}
\caption{Training and test data for robotic/conventional instrument tracking.} 
\label{tab:valdata_track} \end{table*}

\begin{figure}
	\centering
	\subfloat{\includegraphics[width=0.4898\columnwidth]{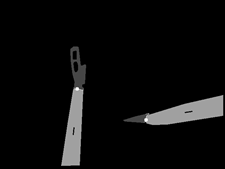}}\hfill
  \subfloat{\includegraphics[width=0.46\columnwidth]{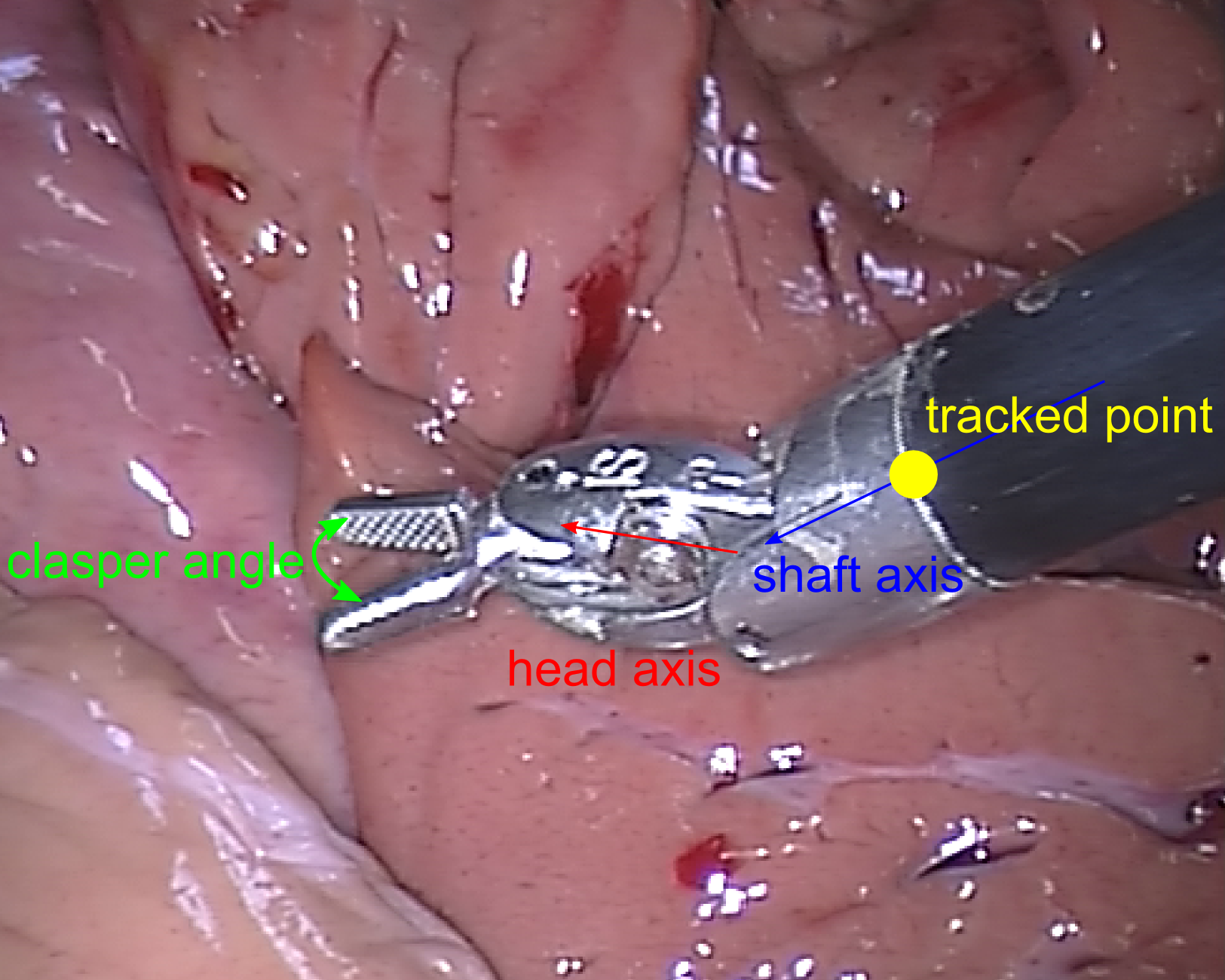}}
	\caption{Example of data provided for instrument tracking. Left shows a mask for a conventional laparoscopic instrument with the point to track in white and the instrument axis in black. Right shows the data provided for the robotic instruments, in yellow the point to track, in red the head axis, in blue the shaft axis and in green the clasper angle.}
	\label{fig:tra-traindata}
\end{figure}

The validation data for tracking originate from the same recordings already described in the previous paragraphs of section \ref{sec:valdata}. In addition to the RGB frames and the two types of annotated masks (\ref{ssec:train_seg}), a csv file with pixel coordinates of the center point and the normalized axis vector of the shaft for each instrument was provided as reference. For the robotic instruments, the normalized axis vector of the instrument head as well as the angle between the claspers is given as well (figure \ref{fig:tra-traindata}). For the test data no reference was provided. The center point is defined as the intersection between the shaft and the metal manipulator on the instrument axis.

For the robotic data set, an annotation is provided for every frame. For the conventional data set, only one frame per second is annotated.

Furthermore, the sequences in the validation data set for both conventional laparoscopic and robotic surgery were categorized according to challenging situations.
For conventional laparoscopic surgery, the sequences contained challenges such as multiple instruments (\DTCMultiple), multiple occurrences of instrument occlusions (\DTCOcclusion), blood (\DTCBlood), smoke (\DTCSmoke) and surgical objects such as meshes and clips (\DTCObjects).
The robotic dataset contained sequences with multiple instruments (\DTRMultiple).
The validation measures were calculated separately for these subsets.

A detailed overview including training and test data is given in table \ref{tab:valdata_track}.
Participants were advised to use the training data in a leave-one-sequence-out fashion. It was not allowed to use the same sequence in the training set when testing the additional frames for each of the sequences (1-4) provided for training. For the new sequences (5-6) the whole training data could be used.

Participants uploaded the pixel coordinates of the center point and axis vector of the shaft for each instrument in each frame in the test dataset. For the robotic dataset, the axis vector of the head as well as the clasper angle was provided as well.

\subsection{Reference method and associated error}

Reference data was provided as annotated masks as described in \ref{ssec:ref_seg} as well as coordinates of specific instrument parts which were generated differently according to the surgical application scenario.

\subsubsection{Robotic laparoscopic instruments}
Using hand corrected kinematics provided by the da Vinci Research Kit (DVRK), the center point, shaft axis, head axis and clasper angle were computed for each frame by projecting their location relative to the 3D model reference frame onto the camera sensor using a pinhole projection, where the camera parameters were computed using the Matlab calibration toolbox. To define the location of the the points in the 3D model reference frame, we manually estimated the coordinates of the 3D tracked point using modeling software and generated the tracked axes and clasper angles from the robot kinematic coordinate frames directly.  

\subsubsection{Conventional laparoscopic instruments}
Given the results provided via crowd-sourcing (\ref{ssec:ref_seg}), the position of the center point  and axis were determined in an automated fashion. The center point is defined as the intersection between the shaft and the metal manipulator on the instrument axis.
First the segmented tip and shaft regions were used to locate the border between shaft and the metal manipulator. We then determined the center and principal axis of the instrument section. Using a line going through the center point with the principal axis as direction, we used the intersection of this line with the border as center point. The principal axis was used as instrument axis.


\subsection{Validation criterion}
\label{ssec:vc:track}
To evaluate the quality of a submitted tracking result, several distance measures were calculated between the reference and the predicted position and axis. To assess the accuracy of the 2D tracked center point, we compute the Euclidean $E_{TP}$ distance between the predicted center point $T_P$ and the ground truth center point $T_{GT}$ for each tool.
$$E_{TP} = ||T_P - T_{GT}||_2$$
We also compute the angular distance $E_{AD}$ between each of the predict angles ($\alpha_P$) for the shaft, wrist and claspers and the groundtruth ($\alpha_{GT}$). 
$$E_{AD} = |\alpha_P - \alpha_{GT}|$$


\section{Results}
\label{sec:results}
\subsection{Instrument segmentation}
\label{sec:results:seg}
For the instrument segmentation, we present two types of results.
First we evaluate the performance of the previously presented methods on both the \DSC\ and the \DSR\ validation sets.
Based on the results of the single methods on these validation sets, we formulate the hypothesis that merging the segmentations of multiple separate methods should provide a measurable improvement.
We propose two different methods for merging different segmentations and evaluate the performance of these on the validation sets.

For each dataset, we then computed the metrics introduced in section \ref{ssec:vc} for all frames and presented the averages for each dataset and their respective subsets. 
The DCS was used to rank the different methods.

To determine the statistical significance of the difference in the DCS between the highest ranking method and each of the lower rankings methods, we performed a Wilcoxon signed-rank test \cite{Wilcoxon45} to compare the distribution of the DSC of two methods on a given dataset.
We tested for a significant difference in DSC values at a $p < 0.01$ significance level, which is coded with a green cell background in the following tables and a $0.01 \le p < 0.05$ significance level, which is represented as a yellow cell background. $p \ge 0.05$ is coded as a red cell background.

\subsubsection{Single results}
\label{sec:results:seg:single}

\paragraph{\DSC}
\label{sec:results:seg:single:dsc}
For the \DSC\ dataset, we present the ranked results of the submitted method on each subset.
In table \ref{tab:results:rigid:single:all} the performances of the methods on the entire test dataset are listed, while table \ref{tab:results:rigid:single:challenges} lists the performances on the different challenge subsets.
\begin{table*}[htb]
\centering
\small
\subfloat[]{
\label{tab:results:rigid:single:all}
\resizebox{!}{1.05cm}{
\begin{tabular}{|l||c|c|c|c|}
\hline
\textbf{\DSCAll}&DSC & Prec. & Rec. & Acc.\\
\hline
SEG-KIT-CNN & 0.88 & 0.86 & 0.90 & 0.98\\
SEG-UB & \cellcolor{green!75}0.84 & 0.78 & 0.94 & 0.97\\
SEG-UCL-CNN & \cellcolor{green!75}0.82 & 0.81 & 0.88 & 0.97\\
SEG-JHU & \cellcolor{green!75}0.82 & 0.83 & 0.85 & 0.97\\
SEG-UGA & \cellcolor{green!75}0.66 & 0.94 & 0.55 & 0.95\\
SEG-KIT-RF & \cellcolor{green!75}0.50 & 0.74 & 0.44 & 0.93\\
SEG-UCL-RF & \cellcolor{green!75}0.42 & 0.74 & 0.35 & 0.93\\
\hline
\end{tabular}
}
}

\subfloat[]{
\label{tab:results:rigid:single:challenges}
\resizebox{!}{1.20cm}{
\begin{tabular}{|l||c|c|c|c||c|c|c|c||c|c|c|c||c|c|c|c|}
\hline
&\multicolumn{4}{c||}{\textbf{\DSCBlood}}&\multicolumn{4}{c||}{\textbf{\DSCMesh}}&\multicolumn{4}{c||}{\textbf{\DSCOverlap}}&\multicolumn{4}{c|}{\textbf{\DSCSmoke}}\\
\cline{2-17}
&DSC & Prec. & Rec. & Acc.&DSC & Prec. & Rec. & Acc.&DSC & Prec. & Rec. & Acc.&DSC & Prec. & Rec. & Acc.\\
\hline
SEG-JHU & \cellcolor{green!75}0.73 & 0.75 & 0.73 & 0.97 & \cellcolor{green!75}0.75 & 0.64 & 0.97 & 0.95 & \cellcolor{red!75}0.86 & 0.87 & 0.86 & 0.96 & \cellcolor{red!75}0.80 & 0.83 & 0.81 & 0.97\\
SEG-KIT-CNN & \textbf{0.85} & 0.87 & 0.84 & 0.98 & \textbf{0.83} & 0.76 & 0.94 & 0.97 & \textbf{0.86} & 0.88 & 0.85 & 0.96 & \textbf{0.86} & 0.86 & 0.87 & 0.98\\
SEG-KIT-RF & \cellcolor{green!75}0.31 & 0.51 & 0.23 & 0.94 & \cellcolor{green!75}0.47 & 0.51 & 0.52 & 0.91 & \cellcolor{green!75}0.52 & 0.82 & 0.42 & 0.89 & \cellcolor{green!75}0.41 & 0.60 & 0.37 & 0.93\\
SEG-UB & \cellcolor{yellow!75}0.76 & 0.66 & 0.94 & 0.96 & \cellcolor{yellow!75}0.76 & 0.66 & 0.96 & 0.95 & \cellcolor{red!75}0.84 & 0.79 & 0.91 & 0.95 & \cellcolor{red!75}0.80 & 0.71 & 0.94 & 0.97\\
SEG-UCL-CNN & \cellcolor{green!75}0.77 & 0.74 & 0.82 & 0.97 & \cellcolor{yellow!75}0.79 & 0.71 & 0.91 & 0.96 & \cellcolor{red!75}0.83 & 0.81 & 0.88 & 0.95 & \cellcolor{green!75}0.77 & 0.74 & 0.85 & 0.96\\
SEG-UCL-RF & \cellcolor{green!75}0.30 & 0.67 & 0.24 & 0.95 & \cellcolor{green!75}0.42 & 0.62 & 0.42 & 0.91 & \cellcolor{green!75}0.46 & 0.83 & 0.37 & 0.88 & \cellcolor{green!75}0.38 & 0.56 & 0.37 & 0.92\\
SEG-UGA & \cellcolor{green!75}0.63 & 0.98 & 0.50 & 0.97 & \cellcolor{red!75}0.68 & 0.80 & 0.68 & 0.93 & \cellcolor{green!75}0.69 & 0.94 & 0.58 & 0.93 & \cellcolor{green!75}0.72 & 0.95 & 0.59 & 0.96\\
\hline
\end{tabular}
}}
\caption{The ranked results of the segmentation methods on all the subsets of \DSC. The statistical significance of the difference in ranking is color-coded: Green indicates a significance value of $p < 0.01$, yellow $p < 0.05$ and red $p \ge 0.05$.}
\label{tab:results:rigid:single}
\end{table*}

It is noticeable that SEG-KIT-CNN outperforms the other methods in all subsets and significantly outperforms the competition on 3 of 5 subsets.
Furthermore a split between the method type can be seen in all subsets, as the top 4 ranks are always taken by the CNN-based methods.
\paragraph{\DSR}
For the \DSR\ dataset, we also present the ranked results of each method (table \ref{tab:results:robotic:single}).
Here SEG-JHU significantly outperforms all the other submitted methods.
While, similarly as for \DSC\ a CNN-based method achieves the highest DCS, a split between CNN-based methods and non CNN-based methods cannot be observed here.
\begin{table}[htb]
\small
\centering
\resizebox{!}{1.05cm}{
\begin{tabular}{|l||c|c|c|c|}
\hline
\textbf{\DSRAll}&DSC & Prec. & Rec. & Acc.\\
\hline
SEG-JHU & 0.88 & 0.84 & 0.92 & 0.97\\
SEG-UCL-CNN & \cellcolor{green!75}0.86 & 0.83 & 0.90 & 0.97\\
SEG-UCL-RF & \cellcolor{green!75}0.85 & 0.87 & 0.83 & 0.96\\
SEG-KIT-CNN & \cellcolor{green!75}0.81 & 0.86 & 0.77 & 0.96\\
SEG-KIT-RF & \cellcolor{green!75}0.78 & 0.86 & 0.72 & 0.95\\
SEG-UGA & \cellcolor{green!75}0.78 & 0.95 & 0.66 & 0.96\\
\hline
\end{tabular}
}
\caption{the ranked results of all segmentation methods on the \DSR\ dataset. The statistical significance of the difference in ranking is color-coded: Green indicates a significance value of $p < 0.01$.}
\label{tab:results:robotic:single}
\end{table}
\subsubsection{Merged results}
While on each dataset one method significantly outperforms the competition (SEG-KIT-CNN on \DSC\ and SEG-JHU on \DSR), we pose the hypothesis that all the information contained in the results of other methods is not necessarily contained in the leading method.
We therefore propose to merge multiple segmentation results in order to determine whether this could improve upon the results of the leading method.

An obvious and naive approach to merge multiple segmentations would be via majority voting (MV).
Here, each method would cast a vote if a given pixel belongs to either the object of interest (a laparoscopic instrument) or the background.
The label that the majority of methods selected is then assigned to the pixel.
The drawback of majority voting is that the vote of each method is weighted equally, not taking into account that the performance or quality of a given segmentation might vary.
In medical image segmentation, the STAPLE algorithm \cite{Warfield04} is often used to merge multiple segmentations from experts and/or algorithms.
STAPLE use an expectation-maximization algorithm to assess the quality of each segmentation according to the spatial distribution of structures and regional homogeneity.
The assessed quality is then used as weight while merging.  
To determine whether a combination of segmentation methods can outperform the highest ranking method, we iterated through all possible combinations of the segmentation results of all methods and compared the results.

In the following sections, we will be presenting the three highest ranked combinations of segmentation methods for each subset.
Furthermore, we calculated the significance of the difference in the DCS of each combination to the highest ranked single method.
To abbreviate the names of the merged methods in the following results, each method is assigned an ID number (table \ref{tab:GroupID}).
\begin{table}[htb]
\centering
\small
\begin{tabular}{l|l}
ID & Method \\
\hline
1 & SEG-JHU\\
2 & SEG-KIT-CNN\\
3 & SEG-KIT-RF\\
4 & SEG-UB\\
5 & SEG-UCL-CNN\\
6 & SEG-UCL-RF\\
7 & SEG-UGA\\
\hline
\end{tabular}
\caption{To simplify the results presented in the following section, we assigned each presented method an ID number.}
\label{tab:GroupID}
\end{table}

\paragraph{\DSC}
In table \ref{tab:results:rigid:merged} we present the ranked results on each subset of the segmentations merged using majority voting.
On \DSCAll, the 3 highest ranking combinations outperform SEG-KIT-CNN significantly, as do the top 2 combinations on \DSCBlood\ and \DSCSmoke.
It is also interesting to note that SEG-KIT-CNN is included in the 3 highest ranking combinations for every subset.
\begin{table*}[htb]
\centering
\small
\subfloat[]{
\resizebox{!}{0.8cm}{
\label{tab:results:rigid:majority}
\begin{tabular}{|l||c|c|c|c|}
\hline
\textbf{\DSCAll}&DSC & Prec. & Rec. & Acc.\\
\hline
2 + 4 + 7 &\cellcolor{green!75}0.89 & 0.90 & 0.89 & 0.98\\
1 + 2 + 4 &\cellcolor{green!75}0.89 & 0.88 & 0.92 & 0.98\\
1 + 2 + 4 + 7 &\cellcolor{green!75}0.88 & 0.87 & 0.92 & 0.98\\
\hline\hline 
SEG-KIT-CNN [2] & 0.88 & 0.86 & 0.90 & 0.98\\
\hline
\end{tabular}
}
}
\subfloat[]{
\label{tab:results:rigid:staple}
\resizebox{!}{0.8cm}{
\begin{tabular}{|l||c|c|c|c|}
\hline
\textbf{\DSCAll}&DSC & Prec. & Rec. & Acc.\\
\hline
2 + 4 + 7 &\cellcolor{green!75}0.89 & 0.90 & 0.89 & 0.98\\
1 + 2 + 4 &\cellcolor{green!75}0.89 & 0.88 & 0.92 & 0.98\\
2 + 4 &\cellcolor{green!75}0.89 & 0.91 & 0.87 & 0.98\\
\hline\hline 
SEG-KIT-CNN [2] & 0.88 & 0.86 & 0.90 & 0.98\\
\hline
\end{tabular}
}
}
\caption{The ranked results of the merged segmentations computed using majority voting \protect\subref{tab:results:rigid:majority} and STAPLE \protect\subref{tab:results:rigid:staple} on all the subsets of \DSC. The statistical significance of the difference in ranking is color-coded: Green indicates a significance value of $p < 0.01$, yellow $p < 0.05$ and red $p \ge 0.05$.}
\label{tab:results:rigid:merged}
\end{table*}

The results of the STAPLE based combination on each subset of \DSC\ is presented in table \ref{tab:results:rigid:staple}.
Similar as with majority voting, SEG-KIT-CNN is also included in the 3 highest ranking combinations for every subset.

\begin{figure}[tb]
	\centering
	\includegraphics[width=\columnwidth]{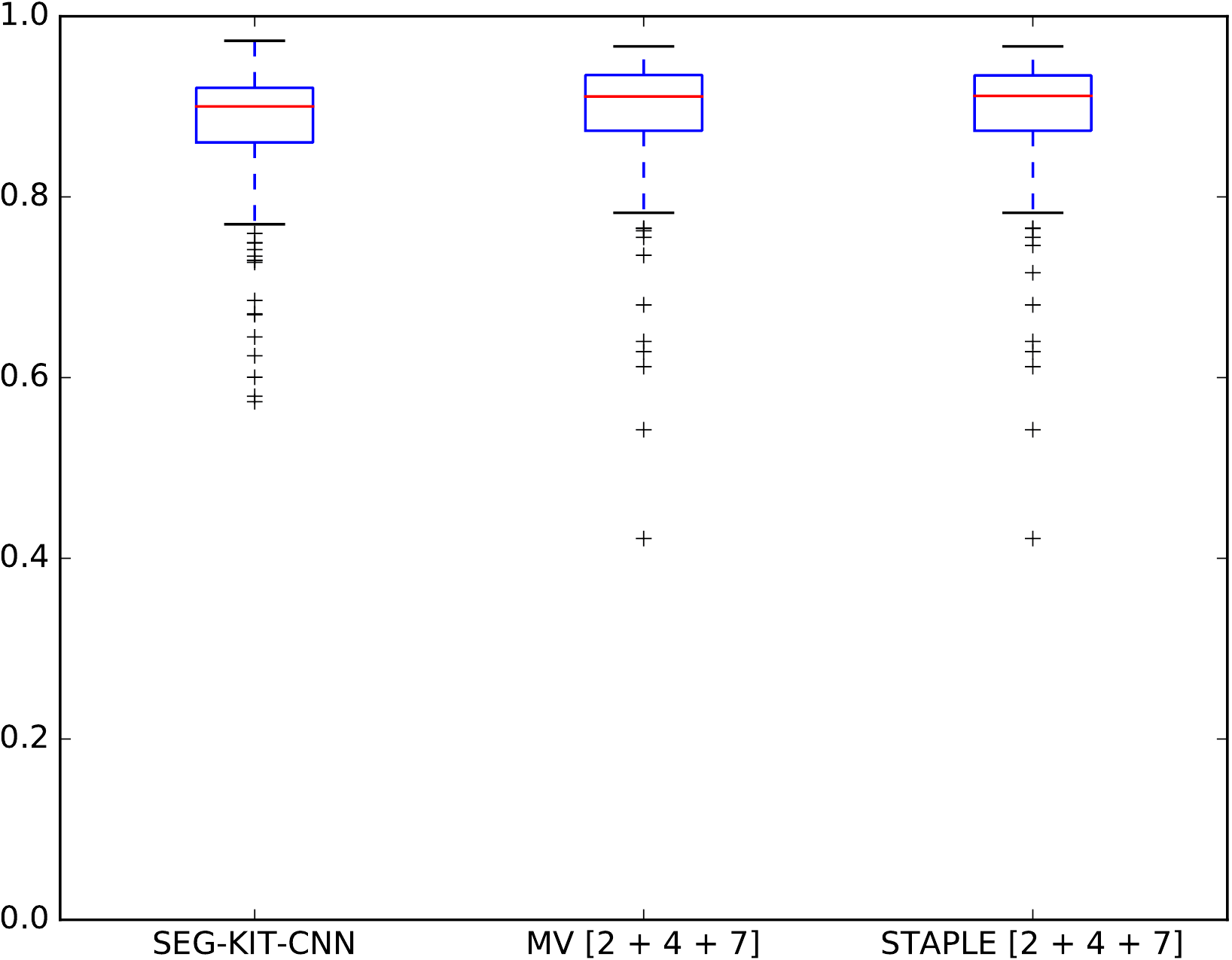}
	\caption{Comparison of all the DSC values and their range of the single and merged methods in \DSCAll.}
	\label{fig:results:rigid:merged}
\end{figure}

In table \ref{tab:results:rigid:merged:comp} and figure \ref{fig:results:rigid:merged}, we compare the performance of the highest ranked single and merged methods.
While both merged methods outperform the highest ranked single method significantly, there is no significant difference in the performance of majority voting and STAPLE (see table \ref{tab:results:rigid:merged:comp}).

\begin{table}[htb]
\centering
\small
\resizebox{!}{0.65cm}{
\begin{tabular}{|l||c|c|c|c|}
\hline
\textbf{\DSCAll}&Mean& SD & Min. & Max.\\
\hline
\cellcolor{red!75}MV[2 + 4 + 7] & 0.89 & 0.08 & 0.42 & 0.97\\
\cellcolor{red!75}STAPLE[2 + 4 + 7] & 0.89 & 0.08 & 0.42 & 0.97\\
\hline\hline
SEG-KIT-CNN & 0.88 & 0.08 & 0.57 & 0.97\\
\hline
\end{tabular}
}
\caption{Comparison of all the DSC values and their range of the single and merged methods on \DSCAll. The color-coded significance here shown is between the results of majority voting and STAPLE. Red indicates a significance value of $p \ge 0.05$.}
\label{tab:results:rigid:merged:comp}
\end{table}

\paragraph{\DSR}
The results for majority voting on \DSR\ can be found in table \ref{tab:results:robotic:majority}.
Here it can be seen that the 3 highest ranked combinations with majority voting outperform the highest ranked single method.
It is also noteworthy that the highest ranked single method on \DSR\ (SEG-JHU) is included in all of the combinations.
\begin{table*}[htb]
\centering
\small
\subfloat[]{
\resizebox{!}{0.8cm}{
\label{tab:results:robotic:majority}
\begin{tabular}{|l||c|c|c|c|}
\hline
\textbf{\DSRAll}&DSC & Prec. & Rec. & Acc.\\
\hline
1 + 4 + 6 &\cellcolor{green!75}0.89 & 0.87 & 0.92 & 0.97\\
1 + 2 + 3 + 4 + 6 &\cellcolor{green!75}0.89 & 0.90 & 0.89 & 0.97\\
1 + 2 + 4 + 6 + 7 &\cellcolor{green!75}0.89 & 0.90 & 0.88 & 0.97\\
\hline\hline
SEG-JHU [1] & 0.88 & 0.84 & 0.92 & 0.97\\
\hline
\end{tabular}
}
}
\subfloat[]{
\label{tab:results:robotic:staple}
\resizebox{!}{0.8cm}{
\begin{tabular}{|l||c|c|c|c|}
\hline
\textbf{\DSRAll}&DSC & Prec. & Rec. & Acc.\\
\hline
1 + 4 + 6 + 7&\cellcolor{green!75}0.89 & 0.87 & 0.92 & 0.97\\
1 + 4 + 6 &\cellcolor{green!75}0.89 & 0.87 & 0.92 & 0.97\\
1 + 2 + 4 + 6 + 7 &\cellcolor{green!75}0.89 & 0.85 & 0.93 & 0.97\\
\hline\hline
SEG-JHU [1] & 0.88 & 0.84 & 0.92 & 0.97\\
\hline
\end{tabular}
}
}
\caption{The ranked results of the merged segmentations computed using majority voting \protect\subref{tab:results:robotic:majority} and STAPLE \protect\subref{tab:results:robotic:staple} on all the subsets of \DSR. The statistical significance of the difference in ranking is color-coded: Green indicates a significance value of $p < 0.01$.}
\end{table*}

In table \ref{tab:results:robotic:staple}, the results for the STAPLE based combinations are listed.
Similarly to majority voting, these combinations also outperform the single methods.

\begin{figure}[tb]
	\centering
	\includegraphics[width=\columnwidth]{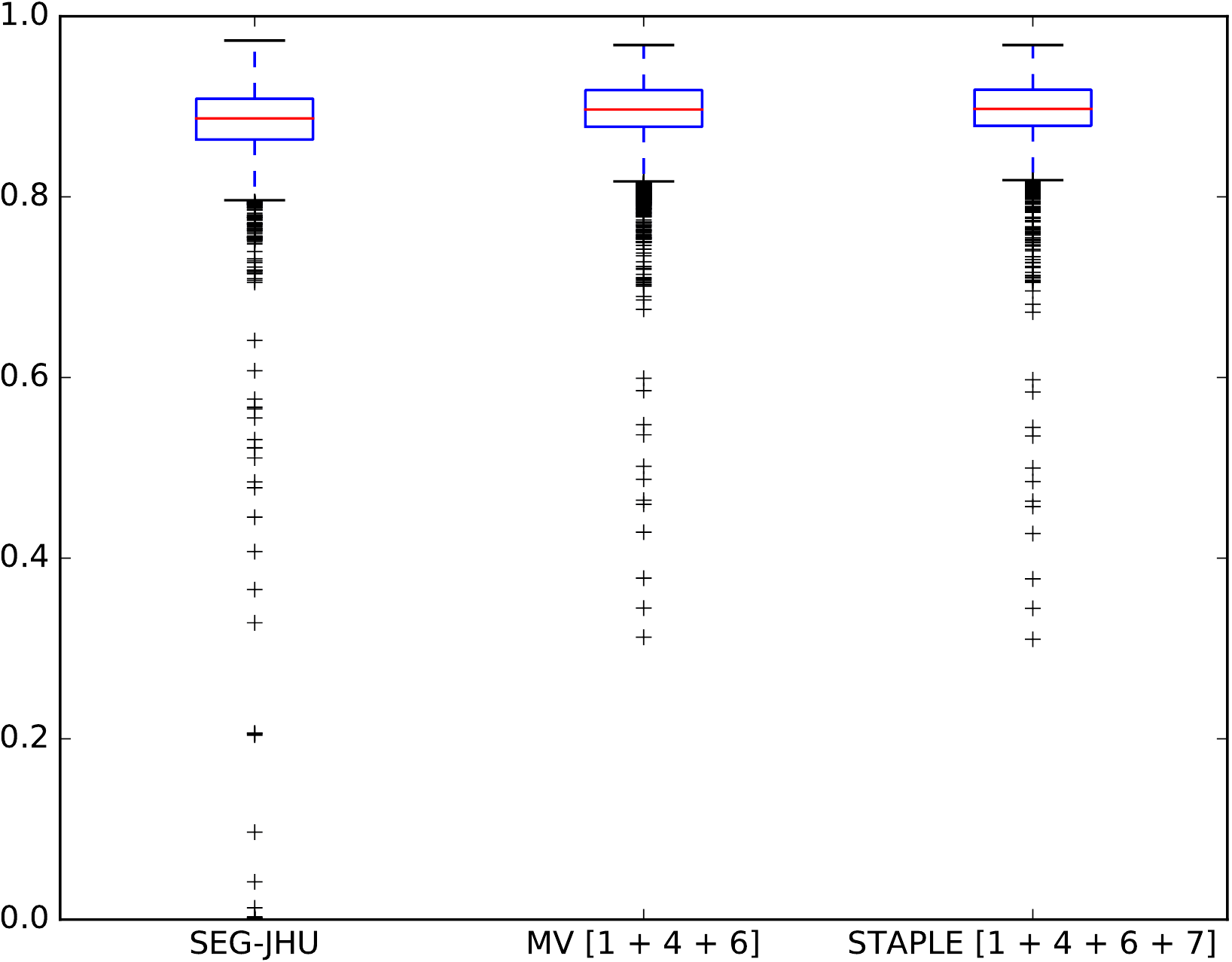}
	\caption{Comparison of all the DSC values and their range of the single and merged methods on \DSR.}
	\label{fig:results:robotic:merged}
\end{figure}

While both majority voting and STAPLE outperform the highest ranked single method, a significant difference between the two combination methods cannot be observed (see table \ref{tab:results:robotic:merged} and figure \ref{fig:results:robotic:merged}).

\begin{table}[htb]
\centering
\small
\resizebox{!}{0.65cm}{
\begin{tabular}{|l||c|c|c|c|}
\hline
\textbf{\DSRAll}&Mean& SD & Min. & Max.\\
\hline
\cellcolor{red!75}MV[1 + 4 + 6] & 0.89 & 0.05 & 0.31 & 0.97\\
\cellcolor{red!75}STAPLE[1 + 4 + 6 +7] & 0.89 & 0.05 & 0.31 & 0.97\\
\hline\hline
SEG-JHU & 0.88 & 0.07 & 0.00 & 0.97\\
\hline
\end{tabular}
}
\caption{Comparison of all the DSC values and their range of single and merged methods on \DSR. The color-coded significance here shown is between the results of majority voting and STAPLE. Red indicates a significance value of $p \ge 0.05$.}
\label{tab:results:robotic:merged}
\end{table}

\subsection{Instrument tracking}
\label{sec:results:tra}

We present results for the instrument tracking by evaluating the performance of all methods on both the \DTC\ and \DTR\ validation sets and, similarly to the segmentation task, testing the hypothesis that a better tracker can be built by combining the tracking output from each method into a single tracker.
We merge the results from the tracker by taking the mean parameter prediction over all methods for each frame.
For the hypothesis that this should improve the tracking accuracy, the errors should be symmetrically distributed around the true value with similar magnitude.
To counter the situation where the errors are not symmetrically distributed, we discard outliers by finding the mean distance between each of measurements for a frame and discarding the largest if it is greater than 2 times the second largest.
The hypothesis here is that we assign higher confidence to the tracking results when there is a greater consensus between the methods.
We again use the Wilcoxon signed-rank test to assess statistical 
significance of the performance changes between each method and use the same color coding as in Section \ref{sec:results:seg}.

\subsubsection{\DTC}
\label{sec:results:tra:single:dtc}

The images in the \DTC\ validation set are highly challenging with complex shadows, numerous occlusions from tissue and out-of-view as well as fast motion (see Figure \ref{fig:seg-valdata}).
Entries to this dataset were made using the TRA-UCL-OL, TRA-KIT and TRA-UGA methods and the results on all the challenge subsets are shown in Table \ref{tab:results:tra:rigid}.
The high tracked point errors are often caused by periods of complete tracking failure, when the method tracks a feature in the background rather than the model tracking the wrong part of the instrument.
Numerous frames in the submissions to the rigid tracking data failed to make any prediction for the instrument position, despite there being an instrument present in the image.
In cases where this happened, we assigned a fixed penalty to the prediction of half the length of the diagonal of the frame for translation and 90$^{\circ}$ for rotation. 

The results show that the merged method achieves the highest accuracy for both the tracked point and the shaft angle.
It is likely that this is caused by the TRA-KIT and TRA-UGA methods providing slight accuracy improvements during frames when TRA-UCL-OL fails to obtain a good estimate of the instrument position and orientation.

\begin{table*}[htb]
\centering
\small
\subfloat[]{
\resizebox{!}{0.75cm}{
\begin{tabular}{|l||c|c|}
\hline
\textbf{\DTCAll}&{T.P. (pix)}&{S.D.}\\
\hline
TRA-MERGED&\cellcolor{white!75}84.7&$11.0^{\circ}$\\
TRA-UCL-OL&\cellcolor{green!75}96.8&\cellcolor{green!75}$17.3^{\circ}$\\
TRA-KIT&\cellcolor{green!75}178.9&\cellcolor{green!75}$19.7^{\circ}$\\
TRA-UGA&\cellcolor{green!75}217.9&\cellcolor{green!75}$41.8^{\circ}$\\
\hline
\end{tabular}
}
}

\subfloat[]{
\resizebox{!}{0.9cm}{
\begin{tabular}{|l||c|c||c|c||c|c||c|c||c|c|}
\hline
&\multicolumn{2}{c||}{\textbf{\DTCBlood}}&\multicolumn{2}{c||}{\textbf{\DTCMultiple}}&\multicolumn{2}{c||}{\textbf{\DTCObjects}}&\multicolumn{2}{c||}{\textbf{\DTCOcclusion}}&\multicolumn{2}{c|}{\textbf{\DTCSmoke}}\\
\cline{2-11}
& T.P. (pix)&S.D.& T.P. (pix)&S.D.& T.P. (pix)&S.D.& T.P. (pix)&S.D.& T.P. (pix)&S.D.\\
\hline
TRA-KIT&\cellcolor{green!75}$233.6$&\cellcolor{green!75}$23.1^{\circ}$&\cellcolor{green!75}$220.9$&\cellcolor{green!75}$24.7^{\circ}$&\cellcolor{green!75}$117.2$&\cellcolor{green!75}$12.8^{\circ}$&\cellcolor{green!75}$225.6$&\cellcolor{green!75}$23.0^{\circ}$&\cellcolor{green!75}$193.9$&\cellcolor{green!75}$23.3^{\circ}$\\
TRA-MERGED&$\mathbf{140.4}$&$15.1^{\circ}$&\cellcolor{green!75}$117.0$&$10.3^{\circ}$&$\mathbf{59.0}$&$8.7^{\circ}$&\cellcolor{green!75}$98.7$&$10.9^{\circ}$&$\mathbf{96.3}$&$13.5^{\circ}$\\
TRA-UCL-OL&\cellcolor{green!75}$181.6$&\cellcolor{green!75}$22.3^{\circ}$&$\mathbf{110.9}$&\cellcolor{green!75}$22.0^{\circ}$&\cellcolor{green!75}$68.3$&\cellcolor{green!75}$15.6^{\circ}$&$\mathbf{87.1}$&\cellcolor{green!75}$20.2^{\circ}$&\cellcolor{green!75}$135.0$&\cellcolor{green!75}$16.1^{\circ}$\\
TRA-UGA&\cellcolor{green!75}$276.4$&\cellcolor{green!75}$55.1^{\circ}$&\cellcolor{green!75}$235.4$&\cellcolor{green!75}$41.0^{\circ}$&\cellcolor{green!75}$228.0$&\cellcolor{green!75}$48.0^{\circ}$&\cellcolor{green!75}$193.8$&\cellcolor{green!75}$33.3^{\circ}$&\cellcolor{green!75}$231.9$&\cellcolor{green!75}$44.9^{\circ}$\\
\hline
\end{tabular}
}
}
\caption{Comparison of the tracking accuracy of each method, averaged across all of \DTC\ datasets. The statistical significance of the difference in ranking is color-coded: Green indicates a significance value of p < 0.01. T.P. refers to the tracked point and S.D. refers to the shaft direction.
\label{tab:results:tra:rigid}}
\end{table*}

\subsubsection{\DTR}
\label{sec:results:tra:single:dtr}

Table \ref{tab:results:tra:robotic} shows the tracking error of each method on the subsets of the \DTR\ dataset for the instrument shaft angle, tracked point and clasper angle.
As no submitted method was capable of tracking the clasper opening angle we omit this degree of freedom from the results.
The results in this tracking task were more balanced, as most methods were able to track for extended sequences and there were few obvious large failures.
This is reflected in the non-improvement of the merged methods.
As in the \DTC\ dataset, TRA-UCL-OL provided the best point tracking but the TRA-UGA method provided the highest shaft angle accuracy.
This is likely due to this method working well only when its quite strict assumptions about the low level image features are not violated, which is regularly not the case during the much more challenging \DTC\ data.
The TRA-UCL-MOD method provided the best wrist direction due to it being the only method which used a full 3D model to perform the tracking.

\begin{table*}[htb]
\centering
\small
\subfloat[]{
\resizebox{!}{0.9cm}{
\begin{tabular}{|l||c|c|c|}
\hline
\textbf{\DTRAll}&T.P. (pix)&S.D.&W.D.\\
\hline
TRA-UCL-OL&$29.7$&$\cellcolor{green!75}6.2^{\circ}$&\cellcolor{green!75}$10.3^{\circ}$\\
TRA-MERGED&\cellcolor{green!75}$32.0$&\cellcolor{green!75}$5.5^{\circ}$&\cellcolor{green!75}$8.8^{\circ}$\\
TRA-UGA&\cellcolor{green!75}$34.9$&$5.2^{\circ}$&\cellcolor{green!75}$13.1^{\circ}$\\
TRA-UCL-MOD&\cellcolor{green!75}$40.2$&\cellcolor{green!75}$8.0^{\circ}$&$6.1^{\circ}$\\
TRA-KIT&\cellcolor{green!75}$106.6$&\cellcolor{green!75}$62.2^{\circ}$&\cellcolor{green!75}$10.4^{\circ}$\\
\hline
\end{tabular}
}
}
\subfloat[]{
\resizebox{!}{0.9cm}{
\begin{tabular}{|l||c|c|c|}
\hline
\textbf{\DTRMultiple}&T.P. (pix)&S.D.&W.D.\\
\hline
TRA-UCL-OL&$38.4$&\cellcolor{green!75}$7.0^{\circ}$&\cellcolor{green!75}$9.4^{\circ}$\\
TRA-MERGED&\cellcolor{green!75}$39.5$&\cellcolor{green!75}$5.7^{\circ}$&\cellcolor{green!75}$7.9^{\circ}$\\
TRA-UGA&\cellcolor{green!75}$40.7$&$5.2^{\circ}$&\cellcolor{green!75}$11.4^{\circ}$\\
TRA-UCL-MOD&\cellcolor{green!75}$45.2$&\cellcolor{green!75}$6.9^{\circ}$&$6.8^{\circ}$\\
TRA-KIT&\cellcolor{green!75}$113.9$&\cellcolor{green!75}$48.3^{\circ}$&\cellcolor{green!75}$12.3^{\circ}$\\
\hline
\end{tabular}
}
}
\caption{Comparison of the tracking accuracy of each method, averaged across all of \DTR\ datasets. The statistical significance of the difference in ranking is color-coded: Green indicates a significance value of p < 0.01. T.P. refers to the tracked point, S.D. refers to the shaft direction and W.D. refers to the wrist direction.
\label{tab:results:tra:robotic}}
\end{table*}
\section{Discussion}
\label{sec:discussion}
\subsection{Instrument Segmentation}
For the instrument segmentation challenge, the methods were evaluated based on the performance on two datasets.
Overall on the \DSC\ dataset, the segmentation method SEG-KIT-CNN achieved the highest performance based on the DSC, while on the \DSR\ dataset SEG-JHU significantly outperformed the other methods.
While the performance of the highest ranked methods was similar on both datasets (DSC of 0.88), the range of the DSC between the sets varied largely, with \DSR\ having a narrower range.
Furthermore, the highest ranked method on each dataset performed significantly worse on the other dataset.

This can be explained by the fact that the \DSC\ dataset was collected from videos of actual laparoscopic operations, while the \DSR\ dataset was collected from ex-vivo organs under controlled conditions.
\DSR\ therefore contains less variance, as only a small selection of surgical instruments were used in the videos and neither the endoscope optics nor the lighting conditions changed.
Also \DSR\ did not include challenges such as smoke and blood.
While the kinematics of the robot allowed automated annotation of the instruments, resulting in more training data for the robotic dataset in contrast to the laparoscopic images that required manual annotation, the similar performance of the highest ranked methods on both datasets seems to suggest that \DSC\ contained a sufficient amount of annotated images.
It is still possible though that the performance of some methods could have improved with more annotated training examples.

It should be noted that on both datasets methods based on CNNs were the highest performing methods.
On \DSC\, all the CNN based methods perform significantly better than the remaining methods, while on \DSR\ one of the RF based methods was among the top three.
Furthermore it can also be seen that on \DSC\, the performance of the RF based methods degraded significantly when confronted with image that contained challenges such as blood or smoke.
Possible explanations for the difference in performance between the CNN based methods and the non-CNN based methods is that, especially the RF based methods, relied on manually selected features that described local pixel neighborhood, while the CNN based methods used features that had been task specifically learned and operated from a more global perspective.
These features allowed the CNN based methods to label each pixel based on information collected from a larger region than its immediate neighbors.
While this allowed for a higher performance, the non-CNN based methods benefit from a run-time closer to real-time than the CNN based methods.
While all methods performed somewhat similarly in precision ($ > 0.70$), the recall performance showed a large spread.
In other words, false positives were rare, instead the methods differed in the amount of correct instrument pixels found.

The ranges of the DSC between the different CNN based methods on \DSC\ (0.82 - 0.88) and on \DSR\ (0.81 - 0.88) are significantly large, prompting the question how these methods differ.
Three of the CNNs (SEG-JHU, SEG-KIT-CNN and SEG-UCL-CNN) are based on FCN, the current state of the art.
While SEG-JHU and SEG-UCL-CNN perform similarly on \DSC, SEG-KIT-CNN outperforms the other two.
This difference between KIT-CNN and the other two lie in weight regularization and using pretrained weights for part of the network, which seems to improve performance, at least on \DSC.
On \DSR, KIT-CNN actually performs significantly worse than the other CNNs.
The method SEG-UB is not based on a FCN, but a patched-based CNN instead.
While this is not the current state of the art, SEG-UB still outperforms two of the FCN-based methods on \DSC.
This can be attributed to the data augmentation used by the method, which none of the other CNN based methods employed.
These results seem to suggest that using a FCN combined with pretraining, data augmentation and regularization increases the DSC.

The results show that merging the segmentation results of different methods improves overall performance.
The merged methods always perform better and the results are often significantly better than best single method.
On both datasets, the highest ranked single methods achieved a DSC of 0.88, while highest ranked merged methods resulted in a DSC of 0.89.
When comparing the manner of merging the results, whether with majority voting or STAPLE, no significant difference was found.

\subsection{Instrument Tracking}
As for the segmentation challenge, two datasets were also used for the instrument tracking challenge for evaluating the performance of the different methods.
On both \DTC\ and \DTR, TRA-UCL-OL outperforms the other single methods significantly in terms of locating the laparoscopic instruments.
The performance of the different methods on \DTC\ differs enormously, which can be contributed to the large variations in the dataset.
On \DTR\ the different methods provided similar performances, except TRA-KIT, whose error was two times larger than the other methods.
This can be contributed to the point, which was tracked. 
TRA-KIT tracked the instrument tip over time, while the other methods tracked the tool center point, which was also the point annotated.

Seeing that TRA-UCL-OL was the only submission based on a machine-learning algorithm, rather than hand-created image processing techniques, leads us to carefully suggest that this seems the more promising direction for future research, though more submissions with more methodological crossover would be required here to draw a more general conclusion.
Tracking the articulation of the tools seems to be a much more challenging problem, as all methods provided errors larger than $10^\circ$ on \DTC\ and larger than $5^\circ$ on \DTR.

The results show that merging the results of the different tracking was beneficial on \DTC\, as the merged method outperformed the single methods in both locating the instrument tip and finding the correct instrument angle.
On \DTR\, merging multiple trackers appears to be less effective, which can be contributed to TRA-KIT tracking a different point than the other methods.

One open problem in comparing tracking results is how to appropriately score, or rather punish, tracking failure.
Ignoring frames in which tracking failed would actually improve performance metrics, while adding a constant error might drastically reduce metrics.

\section{Conclusion}
\label{sec:conclusion}
In this paper, we presented the results on an evaluation of multiple state of the art methods for segmentation and tracking of laparoscopic instruments based on data from the Endoscopic Vision sub-challenge on \textit{Instrument segmentation and tracking}.
For this challenge, a validation data set for two tasks, segmentation and tracking of surgical tools in two settings: robotic and conventional laparoscopic surgery was generated.
Our results indicate that while modern deep learning approaches outperform other methods in instrument segmentation tasks, the results are still not perfect.
Neither did one method clearly have the combined coverage of all the other methods, as merging the segmentation results from different methods achieved a higher performance than any single method alone.

The results from the tracking task show that this is still an open challenge, especially during challenging scenarios in conventional laparoscopic surgery.
Here acquiring more annotated data might be the key to improve results of the machine learning based tracking methods, but acquiring large quantities of training data is challenging.
In the conventional laparoscopic setting, tracking would also be improved by merging results, though this was not the case in the robotic setting, as one method tracked a different part of the instruments than the other methods.

\section*{References}

\bibliography{mybibfile}

\end{document}